\documentstyle[aclap]{article} 
\makeatletter
\def\maketitle{\par
 \begingroup
   \def\thefootnote{\fnsymbol{footnote}}%
   \if@twocolumn
     \twocolumn[\@maketitle]%
     \else \newpage
     \global\@topnum\z@
     \@maketitle \fi\thispagestyle{plain}\@thanks
 \endgroup
 \setcounter{footnote}{0}%
 \let\maketitle\relax
 \let\@maketitle\relax
 \gdef\@thanks{}\gdef\@author{}\gdef\@title{}\let\thanks\relax}
\makeatother
\title{\vspace{-0.5in}Magic for Filter
Optimization in Dynamic Bottom-up Processing}
  \author{Guido Minnen\thanks{\ url:
      http://www.sfs.nphil.uni-tuebingen/sfb/b4}\\
    {\sc sfb} 340, University of T\"ubingen\\Kleine Wilhelmstra\ss e.~113\\
    D-72074 T\"ubingen,\\Germany\\
    {\tt e-mail:minnen@sfs.nphil.uni-tuebingen.de}}
\begin{document}
\maketitle
\vspace{-0.5in}
\begin{abstract}
  Off-line compilation of logic grammars using {\it Magic} allows an
  incorporation of filtering into the logic underlying the grammar.
  The explicit definite clause characterization of filtering resulting
  from Magic compilation allows processor independent and logically
  clean optimizations of {\it dynamic bottom-up processing} with
  respect to goal-directedness.  Two {\it filter optimization}s based
  on the program transformation technique of Unfolding are
  discussed which are of practical and theoretical interest.
\end{abstract}
\section{Introduction}
\marginpar{\vspace*{-13.2cm}\hspace*{3.54cm}\mbox{\scriptsize In:
    Proceedings of the 34th Annual Meeting of the ACL. Santa Cruz, USA. 1996.}}
\label{introduction}
In natural language processing {\it filtering} is used to weed out
those search paths that are redundant, i.e., are not going to be used
in the proof tree corresponding to the natural language expression to
be generated or parsed.  Filter optimization often comprises an
extension of a specific processing strategy such that it exploits
specific knowledge about grammars and/or the computational task(s)
that one is using them for.  At the same time it often remains unclear
how these optimizations relate to each other and what they actually
mean.  In this paper I show how starting from a definite clause
characterization of filtering derived automatically from a logic
grammar using Magic compilation, filter optimizations can be performed
in a processor independent and logically clean fashion.

{\it Magic (templates)} is a general compilation technique for
efficient bottom-up evaluation of logic programs developed in the
deductive database community
\cite{Ramakrishnan:Srivastava:Sudarshan:92}.  Given a logic program,
Magic produces a new program in which the filtering as normally
resulting from top-down evaluation is explicitly characterized
through, so-called, {\it magic predicates}, which produce variable
bindings for filtering when evaluated bottom-up. The original rules of
the program are extended such that these bindings can be made
effective.

As a result of the definite clause characterization of filtering,
Magic brings filtering into the logic underlying the grammar.  I
discuss two filter optimizations.  These optimizations are direction
independent in the sense that they are useful for both generation and
parsing.  For expository reasons, though, they are presented merely on
the basis of examples of generation.

Magic compilation does not limit the information that can be used for
filtering. This can lead to nontermination as the tree fragments
enumerated in bottom-up evaluation of magic compiled grammars are {\it
  connected} \cite{Johnson:forthcoming}. More specifically, 'magic
generation' falls prey to non-termination in the face of head
recursion, i.e., the generation analog of left recursion in parsing.
This necessitates a {\it dynamic} processing strategy, i.e.,
memoization, extended with an abstraction function like, e.g.,
restriction \cite{Shieber:85}, to weaken filtering and a subsumption
check to discard redundant results.  It is shown that for a large
class of grammars the subsumption check which often influences
processing efficiency rather dramatically can be eliminated through
fine-tuning of the magic predicates derived for a particular grammar
after applying an abstraction function in an off-line fashion.

Unfolding can be used to eliminate superfluous filtering steps. Given
an off-line optimization of the order in which the right-hand side
categories in the rules of a logic grammar are processed
\cite{Minnen:Gerdemann:Hinrichs:96} the resulting processing behavior
can be considered a generalization of the head corner generation
approach \cite{Shieber:Vannoord:Moore:Pereira:90}: Without the need to
rely on notions such as {\it semantic head} and {\it chain rule}, a
head corner behavior can be mimicked in a strict bottom-up fashion.
\section{Definite Clause Characterization of Filtering}
Many approaches focus on exploiting specific knowledge about grammars
and/or the computational task(s) that one is using them for by making
filtering explicit and extending the processing strategy such that
this information can be made effective.  In generation, examples of
such extended processing strategies are head corner generation with
its semantic linking \cite{Shieber:Vannoord:Moore:Pereira:90} or
bottom-up (Earley) generation with a semantic
filter~\cite{Shieber:88}.  Even though these approaches often
accomplish considerable improvements with respect to efficiency or
termination behavior, it remains unclear how these optimizations
relate to each other and what comprises the logic behind these
specialized forms of filtering. By bringing filtering into the logic
underlying the grammar it is possible to show in a perspicuous and
logically clean way how and why filtering can be optimized in a
particular fashion and how various approaches relate to each other.
\subsection{Magic Compilation}
Magic makes filtering explicit through characterizing it as definite
clauses.  Intuitively understood, filtering is reversed as binding
information that normally becomes available as a result of top-down
evaluation is derived by bottom-up evaluation of the definite clause
characterization of filtering.
The following is the basic Magic algorithm taken from Ramakrishnan et
al. (1992).
\begin{quote}
Let {\it P} be a program and $q(\overline{c})$ a query on the program.
We construct a new program $P^{mg}$. Initially $P^{mg}$ is empty.
\begin{enumerate}
\item Create a new predicate {\it magic\_p} for each predicate {\it p}
  in {\it P}. The arity is that of {\it p}.
\item For each rule in {\it P}, add the {\it modified version} of the
  rule to $P^{mg}$. If rule {\it r} has head, say, {\it
    p($\overline{t}$)}, the modified version is obtained by adding the
  literal $magic\_p(\overline{t})$ to the body.
\item For each rule {\it r} in {\it P} with head, say, {\it
    p($\overline{t}$)}, and for each literal $q_i(\overline{t_i})$ in
  its body, add a {\it magic rule} to $P^{mg}$. The head is
  $magic\_q_i(\overline{t_i})$. The body contains the literal {\it
    magic\_p($\overline{t}$)}, and all the literals that precede $q_i$ in
  the rule.
\item Create a {\it seed} fact {\it
    magic\_q($\overline{c}$)} from the query.
\end{enumerate}
\end{quote}
To illustrate the algorithm I zoom in on the application of the above
algorithm to one particular grammar rule. Suppose the original grammar
rule looks as follows:

\begin{small}
\begin{tt}
\begin{tabbing}
s(P\=0,P,VForm,SSem):-\\
\>vp(P1,P,VForm,[CSem],SSem),\\
\>np(P0,P1,CSem).
\end{tabbing}
\end{tt}
\end{small}
Step 2 of the algorithm results in the following modified version of
the original grammar rule:

\begin{small}
\begin{tt}
\begin{tabbing}
s(P\=0,P,VForm,SSem):-\\
\>magic\_s(P0,P,VForm,SSem),\\
\>vp(P1,P,VForm,[CSem],SSem),\\
\>np(P0,P1,CSem).
\end{tabbing}
\end{tt}
\end{small}
A magic literal is added to the right-hand side of the rule which
'guards' the application of the rule. This does not change the
semantics of the original grammar as it merely serves as a way to
incorporate the relevant bindings derived with the magic predicates to
avoid redundant applications of a rule.
Corresponding to the first right-hand side literal in the
original rule step 3 derives the following magic rule:

\begin{small}
\begin{tt}
\begin{tabbing}
mag\=ic\_vp(P1,P,VForm,[CSem],SSem):-\\
\>magic\_s(P0,P,VForm,SSem).
\end{tabbing}
\end{tt}
\end{small}
It is used to derive from the guard for the original rule a guard for
the rules defining the first right-hand side literal. The second
right-hand side literal in the original rule leads to the following
magic rule:

\begin{small}
\begin{tt}
\begin{tabbing}
mag\=ic\_np(P0,P1,CSem):-\\
  \>magic\_s(P0,P,VForm,SSem),\\ 
  \>vp(P1,P,VForm,[CSem],SSem).
\end{tabbing}
\end{tt}
\end{small}
Finally, step 4 of the algorithm ensures that a seed is created.
Assuming that the original rule is defining the start category, the
query corresponding to the generation of the {\tt s} ``John buys Mary a
book'' leads to the following seed:

\begin{small}
\begin{tt}
\begin{tabbing}
magic\_s(P0,P,finite,buys(john,a(book),mary)).
\end{tabbing}
\end{tt}
\end{small}
The seed constitutes a representation of the initial bindings provided
by the query that is used by the magic predicates to derive guards.
Note that the creation of the seed can be delayed until run-time,
i.e., the grammar does not need to be recompiled for every possible
query.
\subsection{Example}
\label{example1}
Magic compilation is illustrated on the basis of the simple logic
grammar extract in figure~\ref{fig2}. This grammar has been optimized
automatically for generation \cite{Minnen:Gerdemann:Hinrichs:96}: The
right-hand sides of the rules are reordered such that a simple
left-to-right evaluation order constitutes the optimal evaluation
order.
\begin{figure*}[htbp]
\begin{center}
\begin{small}
\begin{tt}
\begin{tabbing}
(1)\ \=sen\=tence(P0,P,decl(SSem)):-\hspace*{3.4cm}\=(5)\ \=np(\=P0,P,NPSem):-\\
\>\>s(P0,P,finite,SSem).\>\>\>pn(P0,P,NPSem)\\
(2)\>s(P0,P,VForm,SSem):-\>\>(6)\>np(P0,P,NPSem):-\\
\>\>vp(P1,P,VForm,[CSem],SSem).\>\>\>det(P0,P1,NSem,NPSem),\\
\>\>np(P0,P1,CSem),\>\>\>n(P1,P,NSem).\\
(3)\>vp(P0,P,VForm,Args,SSem):-\>\>(7)\>det([a$|$P],P,NSem,a(NSem)).\\
\>\>vp(P0,P1,VForm,[CSem$|$Args],SSem),\>(8)\>v([buys$|$P],P,finite,[I,D,S],buys(S,D,I)).\\
\>\>np(P1,P,CSem).\>(9)\>pn([mary$|$P],P,mary)\\
(4)\>vp(P0,P,VForm,Args,SSem):-\>\>(10)\>n([book$|$P],P,book).\\
\>\>v(P0,P,VForm,Args,SSem).
\end{tabbing}
\end{tt}\end{small}
\end{center}
\caption{{\it Simple head-recursive grammar.}}
\label{fig2}
\end{figure*}
With this grammar a simple top-down generation strategy does not
terminate as a result of the head recursion in rule 3. It is necessary
to use memoization extended with an abstraction function and a
subsumption check. Strict bottom-up generation is not attractive
either as it is extremely inefficient: One is forced to generate all
possible natural language expressions licensed by the grammar and
subsequently check them against the start category. It is possible to
make the process more efficient through excluding specific lexical
entries with a semantic filter. The use of such a semantic filter in
bottom-up evaluation requires the grammar to obey the {\it semantic
  monotonicity constraint} in order to ensure
completeness\cite{Shieber:88} (see below).

The 'magic-compiled grammar' in figure~\ref{fig3} is the result of
applying the algorithm in the previous section to the head-recursive
example grammar and subsequently performing two optimizations
\cite{Beeri:Ramakrishnan:91}: All (calls to) magic predicates
corresponding to lexical entries are removed. Furthermore, data-flow
analysis is used to fine-tune the magic predicates for the specific
processing task at hand, i.e., generation.\footnote{For expository
  reasons some data-flow information that does restrict processing is
  not taken into account. E.g., the fact that the {\tt vp} literal in
  rule 2 is always called with a one-element list is ignored here, but
  see section~\ref{subsumption}.}
\begin{figure*}[tbp]
\begin{center}
\begin{small}
\begin{tt}
\begin{tabbing}
(1)\ \=sen\=tence(P0,P,decl(SSem)):-\hspace*{3.4cm}\=(6)\ \ \=np(\=P0,P,NPSem):-\\
\>\>magic\_sentence(decl(SSem)),\>\>\>magic\_np(NPSem),\\
\>\>s(P0,P,finite,SSem).\>\>\>det(P0,P1,NSem,NPSem),\\
(2)\>s(P0,P,VForm,SSem):-\>\>\>\>n(P1,P,NSem).\\
\>\>magic\_s(VForm,SSem),\>(7)\>det([a$|$P],P,NSem,a(NSem)).\\
\>\>vp(P1,P,VForm,[CSem],SSem),\>(8)\>v([buys$|$P],P,finite,[I,D,S],buys(S,D,I)).\\
\>\>np(P0,P1,CSem).\>(9)\>pn([mary$|$P],P,mary) \\
(3)\>vp(P0,P,VForm,Args,SSem):-\>\>(10)\>n([book$|$P],P,book).\\
\>\>magic\_vp(VForm,SSem),\>(11)\> magic\_s(finite,SSem):-\\
\>\>vp(P0,P1,VForm,[CSem$|$Args],SSem),\>\>\>magic\_sentence(decl(SSem)).\\
\>\>np(P1,P,CSem).\>(12)\> magic\_vp(VForm,SSem):-\\
(4)\>vp(P0,P,VForm,Args,SSem):-\>\>\>\>magic\_s(VForm,SSem).\\
\>\>magic\_vp(VForm,SSem),\>(13)\>magic\_vp(VForm,SSem):-\\
\>\>v(P0,P,VForm,Args,SSem).\>\>\>magic\_vp(VForm,SSem).\\
(5)\>np(P0,P,NPSem):-\>\>(14)\>magic\_np(CSem):-\\
\>\>magic\_np(NPSem),\>\>\>magic\_s(VForm,SSem),\\
\>\>pn(P0,P,NPSem).\>\>\>vp(P1,P,VForm,[CSem],SSem).\\
\>\>\>(15)\>magic\_np(CSem):-\\
\>\>\>\>\>magic\_vp(VForm,SSem),\\
\>\>\>\>\>vp(P0,P1,VForm,[CSem$|$Args],SSem).\\
\end{tabbing}
\end{tt}
\end{small}
\end{center}
\caption{{\it Magic compiled version 1 of the grammar in figure 1.}}
\label{fig3}
\end{figure*}
\begin{figure*}[tbp]
\begin{scriptsize}
\begin{center}
\begin{picture}(500,150)(0,-30)
\put(30,185){'FILTERING TREE'}
\put(250,185){'PROOF TREE'}
\put(0,-40){1.magic\_sentence(decl(buys(j,a(b),m))).}
\put(65,-33){\line(0,0){55}}
\put(0,20){2.magic\_s(finite,buys(j,a(b),m)).}
\put(65,28){\line(-1,2){63}}
\put(65,28){\line(1,1){17}}
\put(80,50){3.magic\_vp(finite,buys(j,a(b),m)).}
\put(120,80){3.magic\_vp(finite,buys(j,a(b),m)).}
\put(90,58){\line(-2,1){37}}
\put(90,58){\line(2,1){37}}
\put(130,88){\line(-3,2){27}}
\put(145,-10){4.vp([buys$|$A],A,finite,[m,a(b),j],buys(j,a(b),m)).}
\put(43,80){5.magic\_np(m).}
\put(330,-10){6.np([m$|$A],A,m).}
\put(200,20){7.vp([buys,m$|$A],A,finite,[a(b),j],buys(j,a(b),m)).}
\put(300,15){\line(2,-1){37}}
\put(300,15){\line(-2,-1){37}}
\put(70,110){8.magic\_np(a(b)).}
\put(396,-10){9.np([a,b$|$A],A,a(b)).}
\put(260,50){10.vp([buys,m,a,b$|$A],A,finite,[j],buys(j,a(b),m)).}
\put(380,45){\line(1,-1){47}}
\put(380,45){\line(-3,-1){47}}
\put(0,160){11.magic\_np(j).}
\put(83,-10){12.np([j$|$A],A,j).}
\put(275,80){13.s([j,buys,m,a,b$|$A],A,finite,buys(j,a(b),m)).}
\put(280,77){\line(-2,-1){155}}
\put(280,77){\line(3,-1){50}}
\put(280,87){\line(0,0){20}}
\put(275,110){15.sentence([j,buys,m,a,b$|$A],A,decl(buys(j,a(b),m))).}
\linethickness{1pt}
\bezier{37}(261,-3)(261,-3)(58,74)
\bezier{43}(331,29)(331,29)(108,105)
\bezier{48}(330,60)(330,60)(16,151)
\end{picture}
\end{center}
\end{scriptsize}
\caption{{\it `Connecting up' facts resulting from semi-naive generation of the
    sentence ``John buys Mary a book'' with the magic-compiled grammar from figure 2.}}
\label{fig4}
\end{figure*}
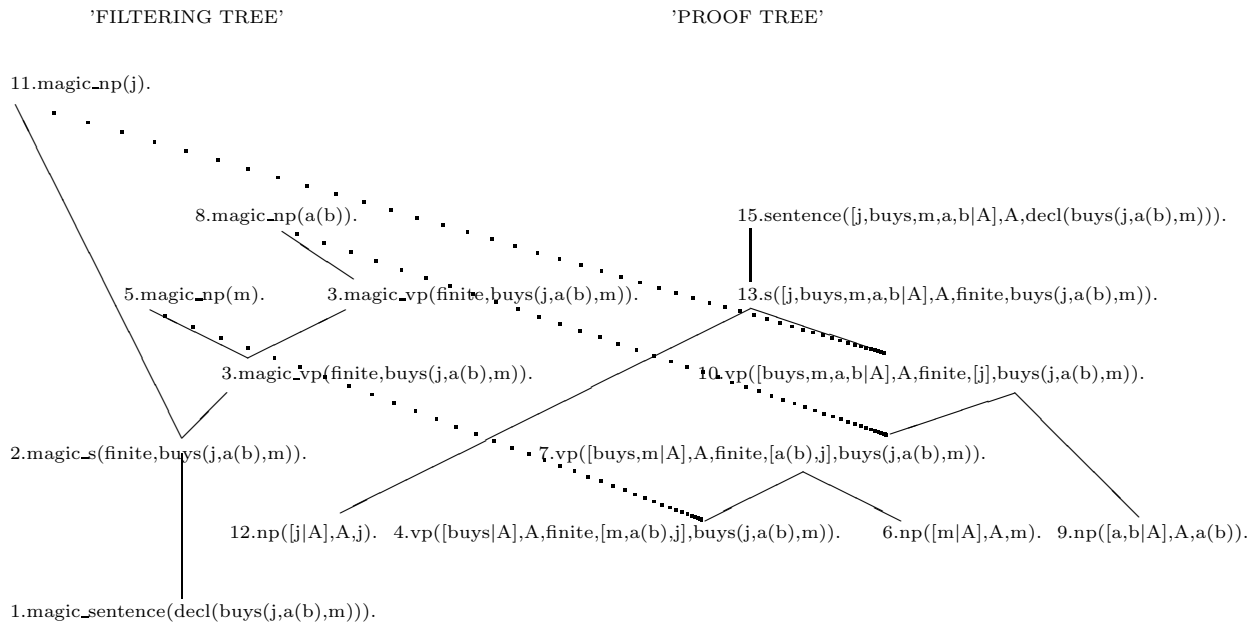
Given a user-specified {\it abstract query}, i.e., a specification of
the intended input \cite{Beeri:Ramakrishnan:91} those arguments which
are not bound and which therefore serve no filtering purpose are
removed.  The modified versions of the original rules in the grammar
are adapted accordingly.  The effect of taking data-flow into account
can be observed by comparing the rules for {\tt magic\_vp} and {\tt
  magic\_np} in the previous section with rule 12 and 14 in figure
\ref{fig3}, respectively.
  
Figure~\ref{fig4} shows the results from generation of the {\tt
  sentence} ``John buys Mary a book''. In the
case of this example the seed looks as follows:

\begin{small}
\begin{tt}
\begin{tabbing}
magic\_sentence(decl(buys(john,a(book),mary))).
\end{tabbing}
\end{tt}
\end{small}
\noindent
The {\it facts}, i.e., passive edges/items, in figure~\ref{fig4}
resulted from {\it semi-naive} bottom-up evaluation
\cite{Ramakrishnan:Srivastava:Sudarshan:92} which constitutes a
dynamic bottom-up evaluation, where repeated derivation of facts from
the same earlier derived facts (as in {\it naive} evaluation;
Bancilhon, 1985) is blocked.  (Active edges are not memoized.)
The figure\footnote{The numbering of the
  facts corresponds to the order in which they are derived. A number
  of lexical entries have been added to the example grammar.  The
  facts corresponding to lexical entries are ignored.  For expository
  reasons the phonology and semantics of lexical entries (except for
  {\tt v}s) are abbreviated by the first letter.  Furthermore the fact
  corresponding to the {\tt vp} ``buys Mary a book John'' is not
  included.} consist of two tree structures (connected through dotted
lines) of which the left one corresponds to the filtering part of the
derivation. The filtering tree is reversed and derives magic facts
starting from the seed in a bottom-up fashion.  The tree on the right
is the proof tree for the example sentence which is built up as a
result of unifying in the derived magic facts when applying a
particular rule. E.g., in order to derive fact 13, magic fact 2
is unified with the magic literal in the modified version of rule 2
(in addition to the facts 12 and 10). This, however, is not
represented in order to keep the figure clear.  Dotted lines are used to
represent when 'normal' facts are combined with magic facts to derive
new magic facts.

As can be reconstructed from the numbering of the facts in
figure~\ref{fig4} the resulting processing behavior is identical to
the behavior that would result from Earley generation as in Gerdemann
(1991) except that the different filtering steps are performed in a
bottom-up fashion. In order to obtain a generator similar to the
bottom-up generator as described in Shieber (1988) the compilation
process can be modified such that only lexical entries are extended
with magic literals. Just like in case of Shieber's bottom-up
generator, bottom-up evaluation of magic-compiled grammars produced
with this Magic variant is only guaranteed to be complete in case the
original grammar obeys the semantic monotonicity constraint.
\section{Filter Optimization through Program Transformation}
As a result of characterizing filtering by a definite clause
representation Magic brings filtering inside of the logic underlying
the grammar.  This allows it to be optimized in a processor
independent and logically clean fashion. I discuss two possible filter
optimizations based on a program transformation technique called {\it
  unfolding} \cite{Tamaki:Sato:84} 
also referred to as partial
execution, e.g., in Pereira and Shieber (1987).  
\subsection{Subsumption Checking} 
\label{subsumption}
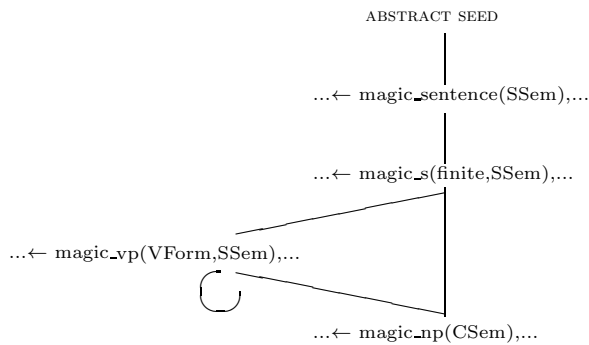
\begin{figure}[h]
\begin{scriptsize}
\begin{center}
\begin{picture}(120,110)(0,-30)
\put(85,80){{\sc abstract seed}}
\put(115,56){\line(-0,-0){19}}
\put(65,50){...$\leftarrow$ magic\_sentence(SSem),...}
\put(115,26){\line(-0,-0){19}}
\put(65,20){...$\leftarrow$ magic\_s(finite,SSem),...}
\put(115,-32){\line(-0,-0){49}}
\put(115,-31){\line(-5,1){79}}
\put(115,15){\line(-5,-1){79}}
\put(-50,-10){...$\leftarrow$ magic\_vp(VForm,SSem),...}
\put(65,-40){...$\leftarrow$ magic\_np(CSem),...}
\put(30,-22.5){\oval(15,15)[tl]}
\put(30,-22.5){\oval(15,15)[bl]}
\put(30,-22.5){\oval(15,15)[br]}
\end{picture}
\end{center}
\end{scriptsize}
\caption{{\it Abstract unfolding tree representing the relation between the
  magic predicates in the compiled grammar.}}
\label{fig5}
\end{figure}

Just like top-down evaluation of the original grammar bottom-up
evaluation of its magic compiled version falls prey to non-termination
in the face of head recursion. It is however possible to eliminate the
subsumption check through fine-tuning the magic predicates derived for
a particular grammar in an off-line fashion.  In order to illustrate
how the magic predicates can be adapted such that the subsumption
check can be eliminated it is necessary to take a closer look at the
relation between the magic predicates and the facts they derive. In
figure~\ref{fig5} the relation between the magic predicates for the
example grammar is represented by an {\it unfolding tree}
\cite{Pettorossi:Proietti:94}. This, however, is not an ordinary
unfolding tree as it is constructed on the basis of an {\it abstract
  seed}, i.e., a seed adorned with a specification of which arguments
are to be considered bound. Note that an abstract seed can be derived
from the user-specified abstract query.  Only the magic part of the
abstract unfolding tree is represented.

The abstract unfolding tree in figure~\ref{fig5} clearly shows why
there exists the need for subsumption checking: Rule 13 in
figure~\ref{fig3} produces infinitely many {\tt magic\_vp} facts.
This 'cyclic' magic rule is derived from the head-recursive {\tt vp}
rule in the example grammar. There is however no reason to keep this
rule in the magic-compiled grammar. It influences neither the
efficiency of processing with the grammar nor the completeness of the
evaluation process. 
\subsubsection{Off-line Abstraction}
\noindent
Finding these types of cycles in the magic part of the compiled
grammar is in general undecidable. It is possible though to `trim' the
magic predicates by applying an abstraction function. As a result of
the explicit representation of filtering we do not need to postpone
abstraction until run-time, but can trim the magic predicates
off-line. One can consider this as bringing abstraction into the logic
as the definite clause representation of filtering is weakened such
that only a mild form of connectedness results which  does not affect
completeness \cite{Shieber:85}.
Consider the following magic rule:

\begin{small}
\begin{tt}
\begin{tabbing}
mag\=ic\_vp(VForm,[CSem$|$Args],SSem):-\\
\>magic\_vp(VForm,Args,SSem).
\end{tabbing}
\end{tt}
\end{small}

\noindent
This is the rule that is derived from the head-recursive {\tt vp} rule
when the partially specified subcategorization list is considered as
filtering information (cf., fn.~1). The rule builds up infinitely
large subcategorization lists of which eventually only one is to be
matched against the subcategorization list of, e.g., the lexical entry
for ``buys''.  Though this rule is not cyclic, it becomes cyclic upon
{\it off-line abstraction}:

\begin{small}
\begin{tt}
\begin{tabbing}
mag\=ic\_vp(VForm,[CSem$|$\_],SSem):-\\
\>magic\_vp(VForm,[CSem2$|$\_],SSem).
\end{tabbing}
\end{tt}
\end{small}
Through trimming this magic rule, e.g., given a bounded term
depth~\cite{Sato:Tamaki:84} or a restrictor~\cite{Shieber:85},
constructing an abstract unfolding tree reveals the fact that a cycle
results from the magic rule. This information can then be used to
discard the culprit.
\subsubsection{Indexing}
\noindent
Removing the direct or indirect cycles from the magic part of the
compiled grammar does eliminate the necessity of subsumption checking
in many cases. However, consider the magic rules 14 and 15 in
figure~\ref{fig3}.  Rule 15 is more general than rule 14. Without
subsumption checking this leads to spurious ambiguity: Both rules
produce a magic fact with which a subject {\tt np} can be built. A
possible solution to this problem is to couple magic rules with the
modified version of the original grammar rule that instigated it. To
accomplish this I propose a technique that can be considered the
off-line variant of an indexing technique described in Gerdemann
(1991).\footnote{This technique resembles an extension of Magic called
  {\it Counting} \cite{Beeri:Ramakrishnan:91}. However, Counting is
  more refined as it allows to distinguish between different levels of
  recursion and serves entirely different purposes.} The indexing
technique is illustrated on the basis of the running example: Rule 14
in figure~\ref{fig2} is coupled to the modified version of the
original {\tt s} rule that instigated it, i.e., rule 2.  Both rules
receive an index:

\begin{small}
\begin{tt}
\begin{tabbing}
s(P0\=,P,VForm,SSem):-\\
\>magic\_s(P0,P,VForm,SSem),\\
\>vp(P1,P,VForm,[CSem],SSem),\\
\>np(P0,P1,CSem,index\_1).\\
\ \\
mag\=ic\_np(CSem,index\_1):-\\
\>magic\_s(P0,P,VForm,SSem),\\
\>vp(P1,P,VForm,[CSem],SSem).
\end{tabbing}
\end{tt}
\end{small}
The modified versions of the rules defining {\tt np}s are adapted such
that they percolate up the index of the guarding magic fact that
licensed its application. This is illustrated on the basis of the
adapted version of rule 14:

\begin{small}
\begin{tt}
\begin{tabbing}
np(\=P0,P,NPSem,INDEX):-\\
\>magic\_np(NPSem,INDEX),\\
\>pn(P0,P,NPSem).
\end{tabbing}
\end{tt}
\end{small}
As is illustrated in section~\ref{example2} this allows the avoidance
of spurious ambiguities in the absence of subsumption check in case of
the example grammar.
\subsection{Redundant Filtering Steps}
Unfolding can also be used to collapse filtering steps.  As becomes
apparent upon closer investigation of the abstract unfolding tree in
figure~\ref{fig5} the magic predicates {\tt magic\_sentence}, {\tt
  magic\_s} and {\tt magic\_vp} provide virtually identical variable
bindings to guard bottom-up application of the modified versions of
the original grammar rules.  Unfolding can be used to reduce the
number of magic facts that are produced during processing. E.g., in
figure~\ref{fig3} the {\tt magic\_s} rule:

\begin{small}
\begin{tt}
\begin{tabbing}
magi\=c\_s(finite,SSem):-\\
\>magic\_sentence(decl(SSem)).\\
\end{tabbing}
\end{tt}
\end{small}
can be eliminated by  unfolding the {\tt magic\_s} literal in
the modified {\tt s} rule:

\begin{small}
\begin{tt}
\begin{tabbing}
s(P0\=,P,VFORM,SSem):-\\
\>magic\_s(VFORM,SSem),\\
\>vp(P1,P,VFORM,,[CSem],SSem),\\
\>np(P0,P1,CSem).\\
\end{tabbing}
\end{tt}
\end{small}
This results in the following new rule which uses the seed for
filtering directly without the need for an intermediate filtering
step:

\begin{small}
\begin{tt}
\begin{tabbing}
\=s(P0\=,P,finite,SSem):-\\
\>\>magic\_sentence(decl(SSem)),\\
\>\>vp(P1,P,finite,[CSem],SSem),\\
\>\>np(P0,P1,CSem).\\
\end{tabbing}
\end{tt}
\end{small}
Note that the unfolding of the {\tt magic\_s} literal leads to the
instantiation of the argument {\tt VFORM} to {\tt finite}. As a result
of the fact that there are no other {\tt magic\_s} literals in the
remainder of the magic-compiled grammar the {\tt magic\_s} rule can be
discarded.

This filter optimization is reminiscent of computing the deterministic
closure over the magic part of a compiled grammar \cite{Doerre:93} at
compile time. Performing this optimization throughout the magic part
of the grammar in figure~\ref{fig3} not only leads to a more succinct
grammar, but brings about a different processing behavior. Generation
with the resulting grammar can be compared best with head corner
generation \cite{Shieber:Vannoord:Moore:Pereira:90} (see next
section).
\subsection{Example}
\label{example2}
\begin{figure*}[t]
\begin{center}
\begin{small}
\begin{tt} 
\begin{tabbing}
(1)\ \ \=sen\=tence(P0,P,decl(SSem)):-\hspace*{3.2cm}\=(6)\ \ \=np(\=P0,P,NPSem,INDEX):-\\
\>\>magic\_sentence(decl(SSem)),\>\>\>magic\_np(NPSem,INDEX),\\
\>\>s(P0,P,finite,SSem).\>\>\>det(P0,P1,NSem,NPSem),\\
(2)\>s(P0,P,finite,SSem):-\>\>\>\>n(P1,P,NSem).\\
\>\>magic\_sentence(decl(SSem)),\>(7)\>det([a$|$P],P,NSem,a(NSem)).\\
\>\>vp(P1,P,finite,[CSem],SSem),\>(8)\>v([buys$|$P],P,finite,[I,D,S],buys(S,D,I)).\\
\>\>np(P0,P1,CSem,index\_1).\>(9)\>pn([mary$|$P],P,mary) \\
(3)\>vp(P0,P,finite,Args,SSem):-\>\>(10)\>n([book$|$P],P,book).\\
\>\>magic\_sentence(decl(SSem)),\>(14)\>magic\_np(CSem,index\_1):-\\
\>\>vp(P0,P1,finite,[CSem$|$Args],SSem),\>\>\>magic\_sentence(decl(SSem)),\\
\>\>np(P1,P,CSem,index\_2).\>\>\>vp(P1,P,finite,[CSem],SSem).\\
(4)\>vp(P0,P,finite,Args,SSem):-\>\>(15)\>magic\_np(CSem,index\_2):-\\
\>\>magic\_sentence(decl(SSem)),\>\>\>magic\_sentence(decl(SSem)),\\
\>\>v(P0,P,finite,Args,SSem).\>\>\>vp(P0,P1,finite,[CSem$|$Args],SSem).\\
(5)\>np(P0,P,NPSem,INDEX):-\\
\>\>magic\_np(NPSem,INDEX),\\
\>\>pn(P0,P,NPSem).
\end{tabbing}
\end{tt}
\end{small}
\end{center}
\caption{{\it Magic compiled version 2 of the grammar in figure 1.}}
\label{fig6}
\end{figure*}
\begin{figure*}[t]
\begin{scriptsize}
\begin{center}
\begin{picture}(500,150)(0,-30)
\put(0,110){11.magic\_np(j,i\_1).}
\put(0,-40){1.magic\_sentence(decl(buys(j,a(b),m))).}
\put(45,-33){\line(-1,6){23}}
\put(45,-33){\line(1,6){14}}
\put(45,-33){\line(2,5){45}}
\put(133,-10){2.vp([buys$|$A],A,finite,[m,a(b),j],buys(j,a(b),m)).}
\put(38,52){3.magic\_np(m,i\_2).}
\put(315,-10){4.np([m$|$A],A,m,i\_2).}
\put(200,20){5.vp([buys,m$|$A],A,finite,[a(b),j],buys(j,a(b),m)).}
\put(300,15){\line(2,-1){37}}
\put(300,15){\line(-2,-1){37}}
\put(52,84){6.magic\_np(a(b),i\_2).}
\put(388,-10){7.np([a,b$|$A],A,a(b),i\_2).}
\put(260,50){10.vp([buys,m,a,b$|$A],A,finite,[j],buys(j,a(b),m)).}
\put(380,45){\line(1,-1){47}}
\put(380,45){\line(-3,-1){47}}
\put(60,-10){11.np([j$|$A],A,j,i\_1).}
\put(275,80){13.s([j,buys,m,a,b$|$A],A,finite,buys(j,a(b),m)).}
\put(280,77){\line(-2,-1){155}}
\put(280,77){\line(3,-1){50}}
\put(280,87){\line(0,0){20}}
\put(275,110){15.sentence([j,buys,m,a,b$|$A],A,decl(buys(j,a(b),m))).}
\linethickness{1pt}
\bezier{37}(261,-3)(261,-3)(63,49)
\bezier{43}(331,29)(331,29)(94,79)
\bezier{48}(330,60)(330,60)(22,104 )
\end{picture}
\end{center}
\end{scriptsize}
\caption{{\it `Connecting up' facts resulting from not-so-naive generation of the
    sentence ``John buys Mary a book'' with the magic-compiled grammar
    from figure 5.}}
\label{fig7}
\end{figure*}
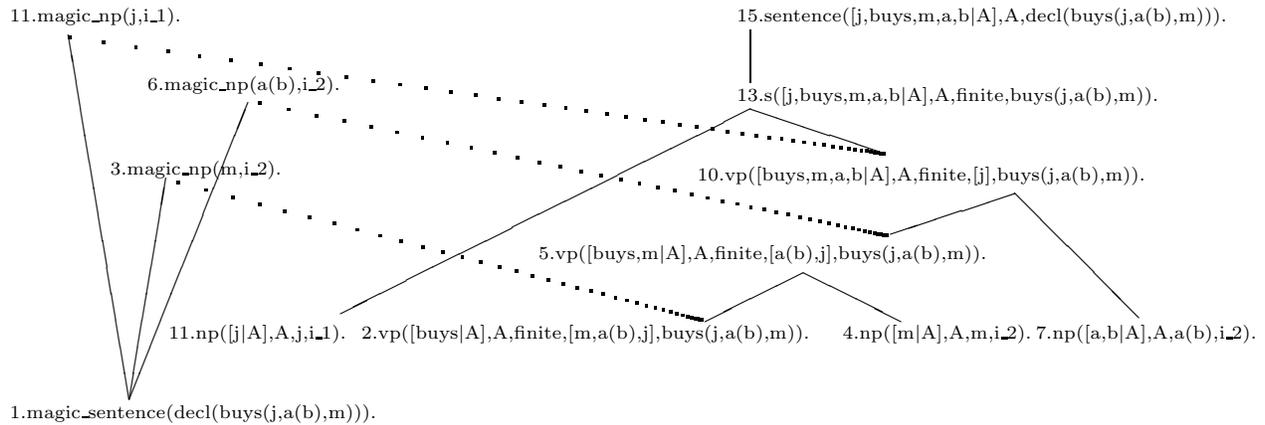

\noindent
After cycle removal, incorporating relevant indexing and the
collapsing of redundant magic predicates the magic-compiled grammar
from figure~\ref{fig3} looks as displayed in figure~\ref{fig6}.
Figure~\ref{fig7} shows the chart resulting from generation of the
sentence ``John buys Mary a book''.\footnote{In addition to the
  conventions already described regarding figure \ref{fig4}, indices
  are abbreviated.} The seed is identical to the one used for the
example in the previous section.  The {\it facts} in the chart
resulted from {\it not-so-naive} bottom-up evaluation: semi-naive
evaluation without subsumption checking
\cite{Ramakrishnan:Srivastava:Sudarshan:92}.
The resulting processing behavior is similar to the behavior that
would result from head corner generation except that the different
filtering steps are performed in a bottom-up fashion.  The head corner
approach jumps top-down from pivot to pivot in order to satisfy its
assumptions concerning the flow of semantic information, i.e.,
semantic chaining, and subsequently generates starting from the
semantic head in a bottom-up fashion. In the example, the seed is used
without any delay to apply the base case of the {\tt vp}-procedure,
thereby jumping over all intermediate chain and non-chain rules. In
this respect the initial reordering of rule 2 which led to rule 2 in
the final grammar in figure~\ref{fig6} is crucial (see
section~\ref{dependency}).
\section{Dependency Constraint on Grammar}
\label{dependency}
To which extent it is useful to collapse magic predicates using
unfolding depends on whether the grammar has been optimized through
reordering the right-hand sides of the rules in the grammar as
discussed in section~\ref{example2}. If the {\tt s} rule in the
running example is not optimized, the resulting processing behavior
would not have fallen out so nicely: In this case it leads either to
an intermediate filtering step for the non-chaining {\tt sentence}
rule or to the addition of the literal corresponding to the subject
{\tt np} to all chain and non-chain rules along the path to the
semantic head. 

Even when cycles are removed from the magic part of a compiled grammar
and indexing is used to avoid spurious ambiguities as discussed in the
previous section, subsumption checking can not always be eliminated.
The grammar must be finitely ambiguous, i.e., fulfill the {\it
  off-line parsability constraint} \cite{Shieber:89}.  Furthermore,
the grammar is required to obey what I refer to as the {\it dependency
  constraint}: When a particular right-hand side literal can not be
evaluated deterministically, the results of its evaluation must
uniquely determine the remainder of the right-hand side of the rule in
which it appears.  Figure~\ref{fig8} gives a schematic example of a
grammar that does not obey the dependency constraint.
\begin{figure}[h]
\begin{center}
\begin{small}
\begin{tt} 
\begin{tabbing}
(1)\ \ \=cat\_1\=(...):-\\
\>\>magic\_cat\_1(Filter),\\
\>\>cat\_2(Filter,Dependency,...),\\
\>\>cat\_3(Dependency).\\
(2)\>magic\_cat\_3(Filter):-\\
\>\>magic\_cat\_1(Filter),\\
\>\>cat\_2(Filter,Dependency,...).\\
\>...\\
(3)\>cat\_2(property\_1,property\_2,...).\\
(4)\>cat\_2(property\_1,property\_2,...).\\
\end{tabbing}
\end{tt}
\end{small}
\end{center}
\caption{{\it Abstract example grammar not obeying the dependency constraint.}}
\label{fig8}
\end{figure}
Given a derived fact or seed {\tt magic\_cat\_1(property\_1)}
bottom-up evaluation of the abstract grammar in figure~\ref{fig8}
leads to spurious ambiguity. There are two possible solutions for {\tt
  cat\_2} as a result of the fact that the filtering resulting from
the magic literal in rule 1 is too unspecific.  This is not
problematic as long as this nondeterminism will eventually disappear,
e.g., by combining these solutions with the solutions to {\tt
  cat\_3}. The problem arises as a result of the fact that these
solutions lead to identical filters for the evaluation of the {\tt
  cat\_3} literal, i.e., the solutions to {\tt cat\_2} do not uniquely
determine {\tt cat\_3}.

Also with respect to the dependency constraint an optimization of the
rules in the grammar is important. Through reordering the right-hand
sides of the rules in the grammar the amount of nondeterminism can be
drastically reduced as shown in Minnen et al.  (1996). This way of
following the intended semantic dependencies the dependency constraint
is satisfied automatically for a large class of grammars.  
\section{Concluding Remarks}
\label{conclusion}
Magic evaluation constitutes an interesting combination of the
advantages of top-down and bottom-up evaluation.  It allows bottom-up
filtering that achieves a goal-directedness which corresponds to
dynamic top-down evaluation with abstraction and subsumption checking.
For a large class of grammars in effect identical operations can be
performed off-line thereby allowing for more efficient processing.
Furthermore, it enables a reduction of the number of edges that need
to be stored through unfolding magic predicates.

\section{Acknowledgments}
The presented research was sponsored by Teilprojekt {\sc b}4 ``From
Constraints to Rules: Efficient Compilation of {\sc hpsg} Grammars'' of the
Sonderforschungsbereich 340 of the Deutsche Forschungsgemeinschaft.
The author wishes to thank Dale Gerdemann, Mark Johnson, Thilo G\"otz
and the anonymous reviewers for valuable comments and discussion.  Of
course, the author is responsible for all remaining errors.

\end{document}